\documentclass[conference]{IEEEtran}
\IEEEoverridecommandlockouts
\usepackage{cite}
\usepackage{amsmath,amssymb,amsfonts}
\usepackage{algorithmic}
\usepackage{graphicx}
\usepackage{textcomp}
\usepackage{xcolor}
\def\BibTeX{{\rm B\kern-.05em{\sc i\kern-.025em b}\kern-.08em
    T\kern-.1667em\lower.7ex\hbox{E}\kern-.125emX}}
\begin{document}

\title{Quantifying and Predicting Residential Building Flexibility Using Machine Learning Methods\\
\thanks{Funding for this work was provided through the Sloan Foundation}
}

\author{\IEEEauthorblockN{Patrick Salter, Qiuhua Huang}
\IEEEauthorblockA{\textit{Department of Electrical Engineering} \\
\textit{Colorado School of Mines}\\
Golden, USA \\
{psalter, qiuhuahuang}@mines.edu}
\and
\IEEEauthorblockN{Paulo Cesar Tabares-Velasco}
\IEEEauthorblockA{\textit{Department of Mechanical Engineering} \\
\textit{Colorado School of Mines}\\
Golden, USA \\
tabares@mines.edu}
}

\maketitle
\thispagestyle{plain} 
\pagestyle{plain}    

\begin{abstract}
Residential buildings account for a significant portion (35\%) of the total electricity consumption in the U.S. as of 2022. As more distributed energy resources are installed in buildings, their potential to provide flexibility to the grid increases. To tap into that flexibility provided by buildings, aggregators or system operators need to quantify and forecast flexibility. Previous works in this area primarily focused on commercial buildings, with little work on residential buildings. To address the gap, this paper first proposes two complementary flexibility metrics (i.e., power and energy flexibility) and then investigates several mainstream machine learning-based models for predicting the time-variant and sporadic flexibility of residential buildings at four-hour and 24-hour forecast horizons. The long-short-term-memory (LSTM) model achieves the best performance and can predict power flexibility for up to 24 hours ahead with the average error around 0.7 kW. However, for energy flexibility, the LSTM model is only successful for loads with consistent operational patterns throughout the year and faces challenges when predicting energy flexibility associated with HVAC systems. 
\end{abstract}

\begin{IEEEkeywords}
Flexibility, residential buildings, machine learning, long short term memory
\end{IEEEkeywords}

\section{Introduction}
Power systems in many countries are facing growing challenges posed by the stochastic and intermittent nature of renewable energy resources. As a result, there has been a push for more flexibility from the demand side of the grid. In the United States, buildings make up the majority of electricity end-use consumption with about 75\% of the nation's electricity consumed in residential and commercial buildings \cite{UseElectricityEnergy}. Buildings usually have some controllable devices and flexible loads and can change consumption patterns to benefit both the grid and the building owners. In order to plan control schemes and respond to requests from the grid, the energy management system (EMS) or controller needs to know how much flexibility the building has and when it is available. Flexibility metrics aim to quantify the flexibility of load and consolidate the various sources of flexibility into a common framework.

Existing work has quantified building load flexibility through a variety of different metrics, but typically the focus is on power, energy, or cost \cite{liEnergyFlexibilityResidential2021}. Power metrics, like those shown in \cite{hirmizPerformanceHeatPump2019} and \cite{caiExperimentalImplementationEmissionaware2023}, aim to quantify the possible power change that could be achieved at a given time. The most prominent of these metrics is peak reduction since reducing consumption during the most congested hours of the day is a well-established topic of research. Because these metrics are simple and just look at power, they are good for predictions further into the future. This simplicity comes at the cost of not providing information on how long the load change can be maintained and what the rebound energy effect of the change will be. Energy metrics, on the other hand, are typically short-term in order to provide this time-dependent information. There are many forms of these metrics, but a common goal is to quantify a building's ability to respond to a demand response event and how efficiently it can do so \cite{renImprovingEnergyFlexibility2021}. Cost metrics, like the flexibility index in \cite{johraInfluenceEnvelopeStructural2019}, are particularly useful for evaluating the cost of a building using its flexibility. These metrics are more evaluative than predictive, mostly used for determining if a load change provides enough benefit to the building owner. There has yet to be a single metric that characterizes all of the pertinent information about load flexibility. As such, for the purpose of forecasting flexibility, at least one power and one energy metric is needed.

Since commercial buildings have a history of EMS, most flexibility forecasting work focuses on these buildings. Compared to residential buildings, commercial buildings have larger thermal masses and tend to have more structured and consistent schedules, making their flexibility easier to predict. HVAC systems \cite{vesaEnergyFlexibilityPrediction2020} and data centers \cite{amasyaliMachineLearningbasedApproach2020} are large flexible loads that have been forecasted to provide services to the grid. There are a variety of forecasting horizons used, but the most common ones are 4 hours and 24 hours. \textbf{Despite the work on commercial buildings, there is little existing work on forecasting of residential building load flexibility} \cite{plaumAggregatedDemandsideEnergy2022}. This is likely due to the fact that individual residential homes typically have relatively small load footprints, limited flexibility potential, and a lack of measurements to enable flexibility quantification and control. However, smart meters and smart home sensors are becoming more common in homes, which provide essential data for building load disaggregation and flexibility quantification. Additionally, building electrification and having more distributed energy resources like energy storage increases their flexibility potential significantly. 

\begin{figure*}[!t]
\centerline{\includegraphics[scale=0.6, trim={0 0 0 0},clip]{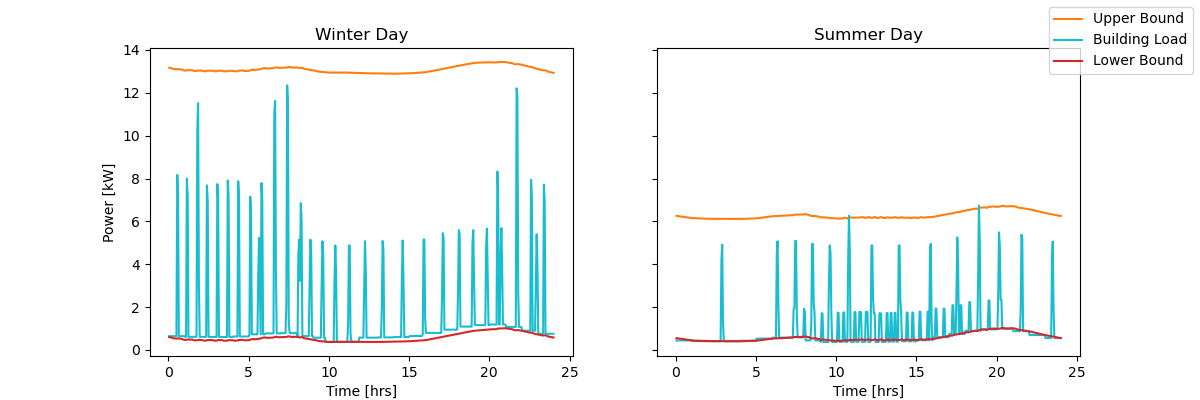}}
\caption{Flexibility bounds for an example summer and winter day}
\label{fig_flexBounds_1}
\end{figure*}

Despite the increased flexibility in residential buildings, there remain several challenges in predicting and using that flexibility. The core of the issue lies in that residential buildings are small loads, have different flexibility sources, and their consumption varies widely based on different occupant habits and preferences. In addition, low thermal mass means the load and flexibility will change much quicker compared to commercial buildings. 
This paper seeks to address this issue and provides data-driven models to forecast the rapidly changing and sporadic flexibility metrics that describe the flexibility offered by a single residential home.

The main contributions of the paper include:
\begin{enumerate}
    \item A framework of metrics to characterize the flexibility of loads within a residential home.
    \item Investigating machine learning models for forecasting these flexibility metrics at horizons of a few hours and a day ahead and confirming their suitability for predicting power flexibility and shortcomings in forecasting energy flexibility. 
\end{enumerate}

The rest of the paper is structured as follows: the flexibility metrics chosen and their calculation methods are described, followed by a discussion on the machine learning methods used for forecasting in Section II. Results and discussions of the models are provided in Section III, followed by conclusions and future work in Section IV.


\section{Flexibility Metrics and Machine Learning Methods for Forecasting}

To create disaggregated load data for training and testing the forecasting model, the building simulation tool EnergyPlus was used with the U.S. Department of Energy's prototype home models in the cool and dry thermal climate zone, which corresponds with Colorado's climate \cite{NRELEnergyPlusEnergyPlus}, \cite{PrototypeBuildingModels}. Although submetering data is uncommon in real-world residential homes, non-intrusive load monitoring techniques can help estimate these disaggregated load profiles. The model used is a manufactured home that has a heat pump for heating and cooling as well as a heat pump water heater. To help the forecasting model learn the temperature dynamics of the house, a dead-zone HVAC controller was implemented based on \cite{cetinDevelopmentValidationHVAC2019}. The water heater in the model already operates in a similar fashion so no changes were made. The building model was simulated for two years with a data resolution of three minutes. For the purposes of this paper, the only sources of flexibility considered are the HVAC and water heater.

\begin{figure}[t]
\centerline{\includegraphics[scale=0.7]{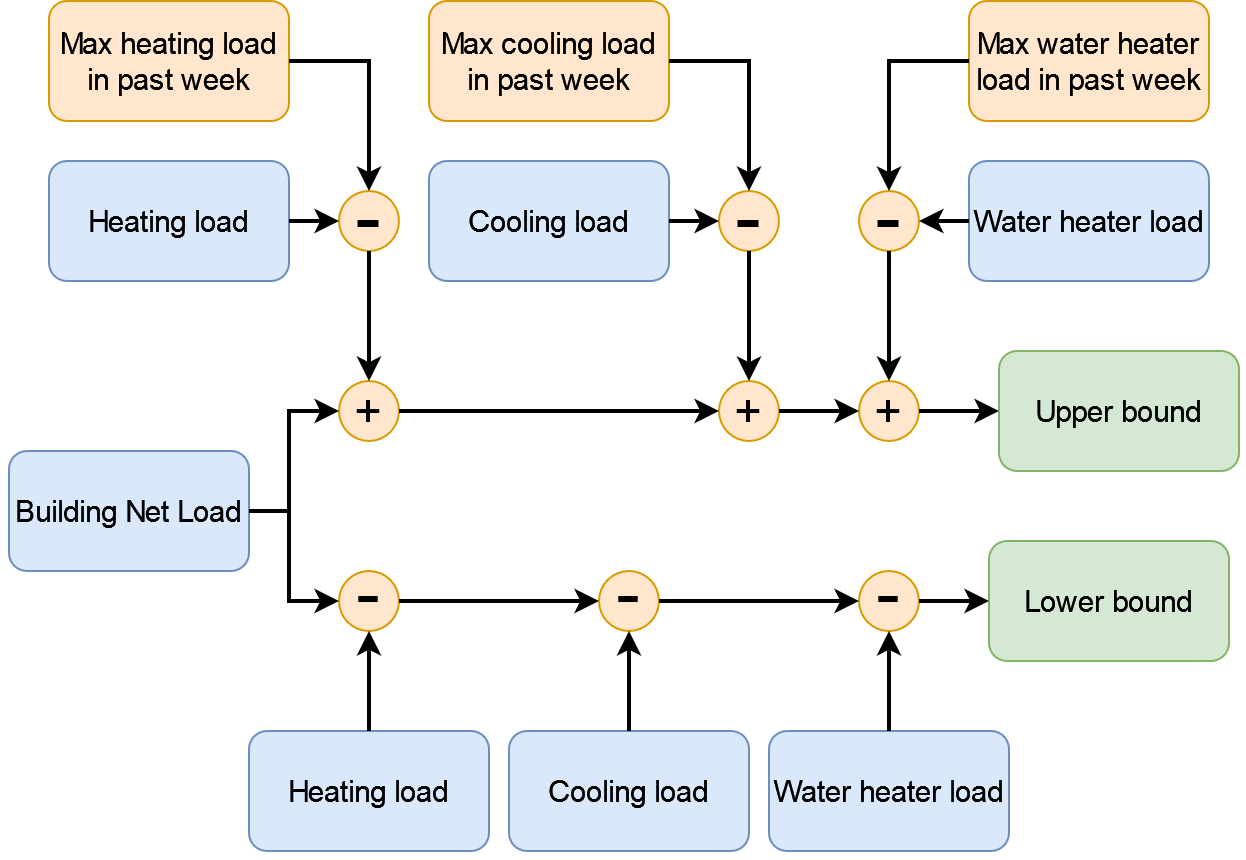}}
\caption{Procedure for calculating the building power flexibility bounds}
\label{fig_flexBounds_2}
\end{figure}

\subsection{Flexibility Metrics}

\begin{figure*}[t]
\centerline{\includegraphics[scale=0.55]{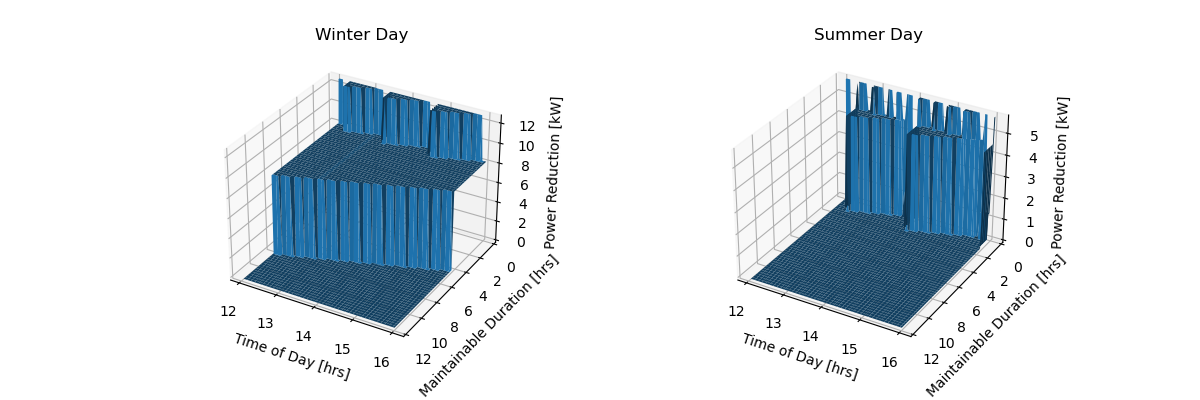}}
\caption{Energy flexibility for an example summer and winter day, 3d plot}
\label{fig_energyFlex_1}
\end{figure*}

\begin{figure}[t]
\centerline{\includegraphics[scale=0.7]{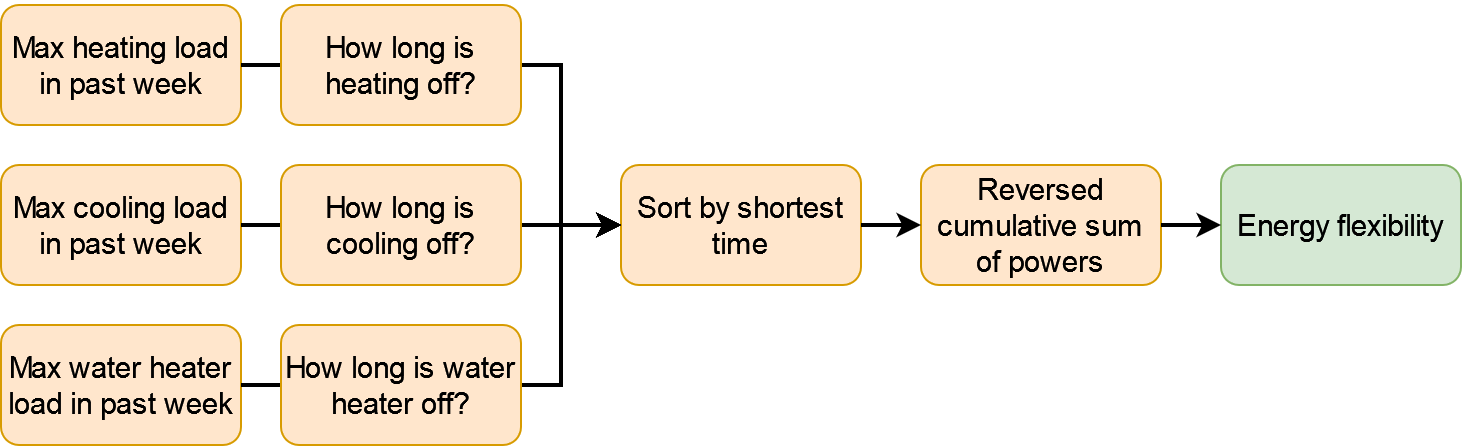}}
\caption{Procedure for calculating the energy flexibility}
\label{fig_energyFlex_3}
\end{figure}

\textbf{Power flexibility bounds} define the upper and lower limit of power consumption. Paired with the expected consumption curve for the building, this provides upward and downward flexibility over a set future horizon. Should a building have the means to support bi-directional power flow, the lower bound has the potential to be negative to represent sending power back to the grid. This metric is calculated using disaggregated load data from the house, as shown in Fig.~\ref{fig_flexBounds_2}. By subtracting the HVAC and water heater electricity consumption from the net load of the building, the lower flexibility bound can be found. As a result, this lower limit is equivalent to the consumption of non-flexible loads in the building. For the upper limit, the difference between rated consumption and current consumption is calculated for the HVAC and water heater and this value is added to the net load. The rated consumption is based on the maximum consumption of HVAC and water heater in the past week. For the HVAC, this value in particular varies depending on the outdoor temperature, so we use a sliding window to allow this behavior to be represented. An example of these bounds are shown in Fig.~\ref{fig_flexBounds_1}. These two limits are highly correlated with one another, both taking on the shape of the non-flexible load consumption with a fairly constant offset between them. Finally, both bounds are smoothed out using a moving average to remove small spikes likely created by the cycling of the house's refrigeration unit. This is done to make training the models easier, however it does come with the consequence that the consumption of the house will occasionally exceed the bounds slightly. This results in the bounds slightly underestimating their limits.

\textbf{Energy flexibility} aims to pair a possible load change in power with a duration that the change can be maintained without violating operating or comfort constraints. These constraints consist of occupant preferences like the acceptable range of indoor air temperatures, or are based on safety like the temperature limits required to avoid unwanted bacteria growth in a hot water tank. For a single house with electric HVAC and water heating, this metric will take the form of an irregular descending staircase for each timestep, as shown in Fig.~\ref{fig_energyFlex_1}. This discontinuity combined with the fast changes makes forecasting energy flexibility a very challenging task. The number of power thresholds relies entirely on the amount of flexible loads in the house, but combining the energy flexibility of several homes at once allows for more granular power reduction and duration offerings. Similar to the flexibility bounds, this metric relies on the disaggregated load profiles of the house. The potential power reduction and maintainable duration are calculated separately for each flexible load. The power thresholds are based on the rated consumption for each device, which is the maximum consumption in the past week, same as the flexibility bounds. The maintainable duration is essentially the amount of time it will take for the indoor air temperature or hot water tank temperature to exceed the constraints without any effort from their respective appliance. The deadband controllers are used to create this information. When the appliances are off, the indoor air and hot water tank temperatures change based on the natural thermal dynamics of the house, providing data on how the temperatures change based on the current state of the house and environment. The power and duration for each devices are paired together and sorted based on duration, from shortest to longest. Finally, a reverse cumulative sum of the powers is used to create the different power tiers.

\subsection{Time Series Forecasting}

The EnergyPlus simulation data was post-processed using the methods detailed in the previous section to calculate the two flexibility metrics. 70\% of the data was used for training, with validation and testing making up the remaining 20\% and 10\%, respectively. Four different models were tested for the forecasting of the flexibility metrics: a linear model, an artificial neural network (ANN), a convolution neural network (CNN), and a recurrent neural network (RNN) using long short-term memory cells (LSTM)\cite{goodfellow2016deep}. The parameters for these models are listed in Table~\ref{tab1}. Different numbers of layers and neurons were tested, and these parameters were selected as they achieved the desired amount of accuracy wihout overfitting to the training data. For the power flexibility bounds, a single model with two outputs was used, one for the upper bound and one for the lower bound. From the simulation data, the net load of the house, outdoor and indoor air temperature, and hot water tank temperature are all used as inputs to the model. Also included as input is the most recent operating state of the HVAC. Heating and cooling have different consumption patterns, which has a large impact on the bounds, particularly the upper bound. Additionally, a fourier analysis of the bounds showed a strong daily periodicity, so a time of day input was also included. The goal for this metric is to achieve an accurate prediction 24 hours ahead.

\begin{table}[!t]
\caption{Forecasting model parameters}
\begin{center}
\begin{tabular}{|c|c|c|}
\hline
&\multicolumn{2}{|c|}{\textbf{Model Parameters}} \\
\cline{2-3} 
\textbf{Model} & \textbf{\textit{Number of layers/neurons}}& \textbf{\textit{Notes}} \\
\hline
Linear& 1 layer with a neuron per input & No activation function\\
\hline
ANN& 2 layers with 512 neurons &  \\
\hline
CNN& 1 layer with 512 neurons & Kernel size of 3 \\
\hline
LSTM& 1 layer with 32 cells &  \\
\hline
\end{tabular}
\label{tab1}
\end{center}
\end{table}

For energy flexibility, since the metric is discontinuous in power and time duration (see Fig. \ref{fig_energyFlex_1}), they are predicted separately to simplify the problem. The power reduction can be readily calculated from past meter data, so the focus was on forecasting the duration the power reduction could be maintained. Predicting this for both HVAC and water heater in a single model proved difficult, so three models were used, one for heating, one for cooling, and one for the water heater. These models all use data on the outdoor air temperature and a time input with a daily periodicity. The water heater model also uses the hot water tank temperature and a binary variable for the state of the water heater. Similarly, the heating and cooling models use indoor air temperature and a binary variable for the HVAC's heating or cooling state. The goal of this metric is to achieve an accurate prediction four hours ahead.

\section{Results and Discussions}
We trained the four machine learning models mentioned above for predicting power flexibility and energy flexibility on a rolling horizon (depending on the prediction horizon) for two years. Thus, each single model covers all four seasons. 
\subsection{Power Flexibility Bounds}

\begin{figure}[t]
\centerline{\includegraphics[scale=0.45, trim={0.75cm 0.25cm 1.5cm 0.7cm},clip]{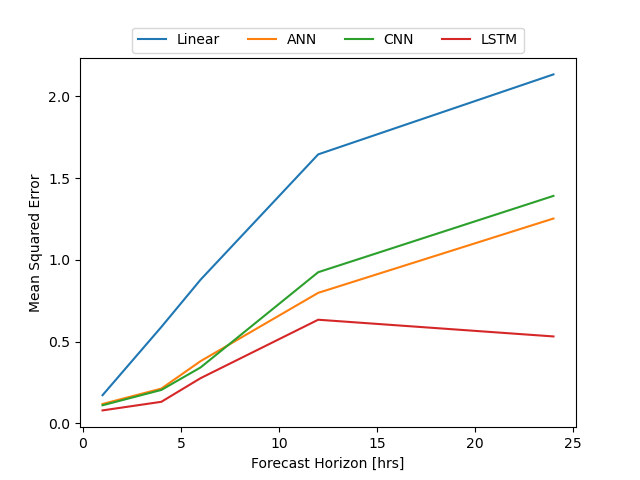}}
\caption{Forecasting error for flexibility bounds over various forecast horizons}
\label{fig_results_flexBounds_3}
\end{figure}

\begin{figure}[t]
\centerline{\includegraphics[scale=0.45, trim={0 0.25cm 1.5cm 0.7cm},clip]{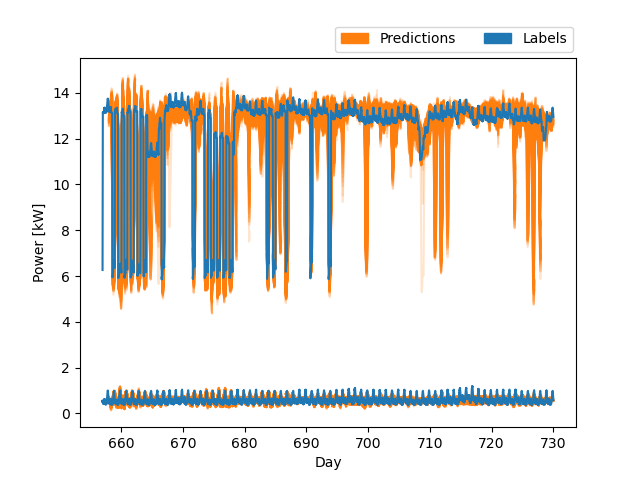}}
\caption{Predicted flexibility bounds using the LSTM model with a forecast horizon of 24 hours}
\label{fig_results_flexBounds_2}
\end{figure}

Fig.~\ref{fig_results_flexBounds_3} shows the mean squared error (MSE) for the four different models. As might be expected, the LSTM model performs the best, achieving an MSE of about 0.5, which is around 0.7 kW. This model performs the best throughout the various forecast horizons, still providing reasonably accurate predictions with a forecast horizon of 24 hours. This reaches the goal of providing information on flexibility one day ahead.

The primary limitation of the model can be seen in Fig.~\ref{fig_results_flexBounds_2} in how it struggles to capture the large change that occurs in the upper bound when switching between heating and cooling. The prediction tends to lag behind the label when these large changes occur and the model has a habit of over-exaggerating the change. Knowing the model's tendency to overshoot allows for these shortcomings to be worked around since they still follow the overall shape of the label and tend to yield conservative results. Another issue is when the model predicts that the upper bound will decrease as if the HVAC is cooling the house despite that not occuring. The tail end of the data in Fig.~\ref{fig_results_flexBounds_2} shows this problem quite well with maximum errors of about 6 kW. These errors increase in frequency as the forecast horizon lengthens. Using separate models for different periods (or seasons) of the year would likely help reduce these errors. 

\subsection{Energy Flexibility}
\begin{figure}[ht]
\centerline{\includegraphics[scale=0.45, trim={0 0.25cm 1.5cm 0.7cm},clip]{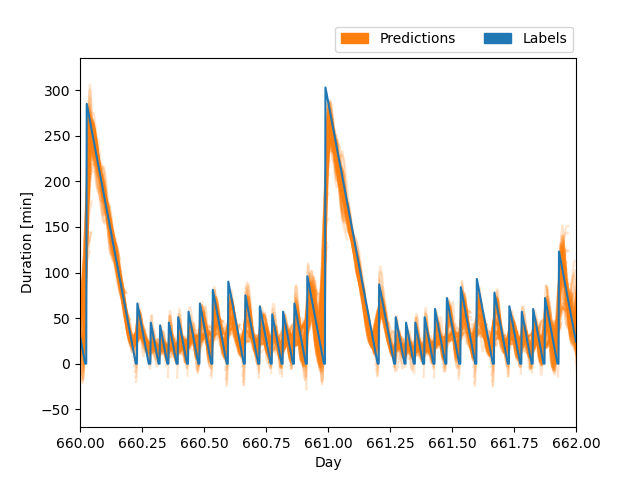}}
\caption{Predicted maintainable duration for water heater with a forecast horizon of four hours}
\label{fig_results_energyFlex_WH_1}
\end{figure}

The model for energy flexibility was split up based on the source of flexibility. For the water heater model, LSTM is the most accurate model. It performs best with a two-hour forecast horizon, achieving an MSE of 384, which is about 20 minutes. With a four-hour horizon, the model shows a noticeable bias towards undershooting the duration. The overall accuracy doesn't significantly decrease, still achieving an average error of around 20 minutes. Several test days can be seen in Fig.~\ref{fig_results_energyFlex_WH_1}.

\begin{figure}[!t]
\centerline{\includegraphics[scale=0.45, trim={0.25cm 0.25cm 1.5cm 0.7cm},clip]{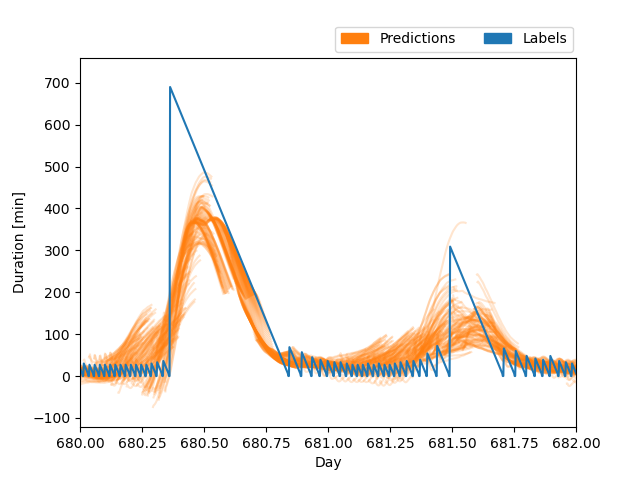}}
\caption{Predicted maintainable duration for HVAC heating with a forecast horizon of four hours}
\label{fig_results_energyFlex_heating_1}
\end{figure}

Even with separate models for heating and cooling, none of the models are capable of accurately forecasting the energy flexibility duration over the year, as seen in Fig.~\ref{fig_results_energyFlex_heating_1}. 
Our analysis showed that this is because of periods of time where the indoor air temperature changes based solely on external factors like the outdoor temperature and solar radiation. The temperature inside may rise even though there is no heating effort from the HVAC. These periods of time can last for multiple hours and create large spikes in the duration, making forecasting quite difficult. These large spikes also are not outliers because they happen frequently, so removing them from the dataset is nontrivial. In light of this, we believe that this method is insufficient for forecasting HVAC flexibility. The consumption of HVAC is much less consistent throughout the year compared to the water heater, likely due to the fact that the hot water tank is located inside the house and does not experience significant changes in the air temperature around it. The goal of the heating and cooling models is to predict the amount of time it will take for the air temperature inside to exceed the comfort limits of the deadband without any effort from the HVAC system. Yet, the machine learning-based forecasting models considered in this paper do not explicitly capture these thermal dynamics which are important for flexibility duration prediction. One possible next step would be to consider a gray box model that incorporates more physics. In particular, an RC model to represent the thermal dynamics of a house seems promising so long as there is an effective method to calibrate the model to a specific house. 

\section{Conclusions and future work}
This paper proposes power flexibility and energy flexibility as two complementary metrics for quantifying residential building load flexibility and develops machine learning methods for forecasting this flexibility. Day-ahead forecasting of power flexibility is viable using an LSTM model, and future work on this metric will look into how to effectively split the year into multiple models to achieve higher accuracy. Despite the success in forecasting energy flexibility for water heaters, our study showed that energy flexibility forecasting is a generally much more challenging task for mainstream machine-learning methods due to the highly non-linear, time-variant, and sporadic nature of individual home energy flexibility. Considering that air temperature dynamics and HVAC system operation modes appear to be difficult for these models to learn purely from data, a continuation of this work will look into gray-box models that can leverage physics to make up for the shortcomings in purely data-driven methods. Another key aspect of flexibility that was outside the scope of this paper is energy storage. Batteries are becoming more common in residential homes and present a significant amount of potential for load flexibility.
\bibliographystyle{IEEEtran}
\bibliography{bibliography}

\begin{thebibliography}{10}
\providecommand{\url}[1]{#1}
\csname url@samestyle\endcsname
\providecommand{\newblock}{\relax}
\providecommand{\bibinfo}[2]{#2}
\providecommand{\BIBentrySTDinterwordspacing}{\spaceskip=0pt\relax}
\providecommand{\BIBentryALTinterwordstretchfactor}{4}
\providecommand{\BIBentryALTinterwordspacing}{\spaceskip=\fontdimen2\font plus
\BIBentryALTinterwordstretchfactor\fontdimen3\font minus \fontdimen4\font\relax}
\providecommand{\BIBforeignlanguage}[2]{{%
\expandafter\ifx\csname l@#1\endcsname\relax
\typeout{** WARNING: IEEEtran.bst: No hyphenation pattern has been}%
\typeout{** loaded for the language `#1'. Using the pattern for}%
\typeout{** the default language instead.}%
\else
\language=\csname l@#1\endcsname
\fi
#2}}
\providecommand{\BIBdecl}{\relax}
\BIBdecl

\bibitem{UseElectricityEnergy}
``Use of electricity - {{U}}.{{S}}. {{Energy Information Administration}} ({{EIA}}),'' https://www.eia.gov/energyexplained/electricity/use-of-electricity.php.

\bibitem{liEnergyFlexibilityResidential2021}
H.~Li, Z.~Wang, T.~Hong, and M.~A. Piette, ``Energy flexibility of residential buildings: {{A}} systematic review of characterization and quantification methods and applications,'' \emph{Advances in Applied Energy}, vol.~3, p. 100054, Aug. 2021.

\bibitem{hirmizPerformanceHeatPump2019}
R.~Hirmiz, H.~M. Teamah, M.~F. Lightstone, and J.~S. Cotton, ``Performance of heat pump integrated phase change material thermal storage for electric load shifting in building demand side management,'' \emph{Energy and Buildings}, vol. 190, pp. 103--118, May 2019.

\bibitem{caiExperimentalImplementationEmissionaware2023}
H.~Cai and P.~Heer, ``Experimental implementation of an emission-aware prosumer with online flexibility quantification and provision,'' Oct. 2023.

\bibitem{renImprovingEnergyFlexibility2021}
H.~Ren, Y.~Sun, A.~K. Albdoor, V.~V. Tyagi, A.~K. Pandey, and Z.~Ma, ``Improving energy flexibility of a net-zero energy house using a solar-assisted air conditioning system with thermal energy storage and demand-side management,'' \emph{Applied Energy}, vol. 285, p. 116433, Mar. 2021.

\bibitem{johraInfluenceEnvelopeStructural2019}
H.~Johra, P.~Heiselberg, and J.~L. Dr{\'e}au, ``Influence of envelope, structural thermal mass and indoor content on the building heating energy flexibility,'' \emph{Energy and Buildings}, vol. 183, pp. 325--339, Jan. 2019.

\bibitem{vesaEnergyFlexibilityPrediction2020}
A.~V. Vesa, T.~Cioara, I.~Anghel, M.~Antal, C.~Pop, B.~Iancu, I.~Salomie, and V.~T. Dadarlat, ``Energy {{Flexibility Prediction}} for {{Data Center Engagement}} in {{Demand Response Programs}},'' \emph{Sustainability}, vol.~12, no.~4, p. 1417, Jan. 2020.

\bibitem{amasyaliMachineLearningbasedApproach2020}
K.~Amasyali, M.~Olama, and A.~Perumalla, ``A {{Machine Learning-based Approach}} to {{Predict}} the {{Aggregate Flexibility}} of {{HVAC Systems}},'' in \emph{2020 {{IEEE Power}} \& {{Energy Society Innovative Smart Grid Technologies Conference}} ({{ISGT}})}, Feb. 2020, pp. 1--5.

\bibitem{plaumAggregatedDemandsideEnergy2022}
F.~Plaum, R.~Ahmadiahangar, A.~Rosin, and J.~Kilter, ``Aggregated demand-side energy flexibility: {{A}} comprehensive review on characterization, forecasting and market prospects,'' \emph{Energy Reports}, vol.~8, pp. 9344--9362, Nov. 2022.

\bibitem{NRELEnergyPlusEnergyPlus}
``{{NREL}}/{{EnergyPlus}}: {{EnergyPlus}}\texttrademark{} is a whole building energy simulation program that engineers, architects, and researchers use to model both energy consumption and water use in buildings.'' https://github.com/NREL/EnergyPlus.

\bibitem{PrototypeBuildingModels}
``Prototype {{Building Models}} | {{Building Energy Codes Program}},'' https://www.energycodes.gov/prototype-building-models.

\bibitem{cetinDevelopmentValidationHVAC2019}
K.~S. Cetin, M.~H. Fathollahzadeh, N.~Kunwar, H.~Do, and P.~C. {Tabares-Velasco}, ``Development and validation of an {{HVAC}} on/off controller in {{EnergyPlus}} for energy simulation of residential and small commercial buildings,'' \emph{Energy and Buildings}, vol. 183, pp. 467--483, Jan. 2019.

\bibitem{goodfellow2016deep}
I.~Goodfellow, Y.~Bengio, and A.~Courville, \emph{Deep learning}.\hskip 1em plus 0.5em minus 0.4em\relax MIT press, 2016.

\end{thebibliography}

\end{document}